\pdfoutput=1

\documentclass[11pt]{article}

\usepackage{ACL2023}

\usepackage{times}
\usepackage{latexsym}

\usepackage[T1] {fontenc}

\usepackage[utf8]{inputenc}

\usepackage{microtype}

\title{Instructions for *ACL Proceedings}

\usepackage{times}
\usepackage{latexsym}
\usepackage{soul}
\usepackage{amsmath}
\usepackage{booktabs}
\usepackage{bm}
\usepackage{cleveref}
\usepackage{graphicx}
\usepackage{tabularx}
\usepackage{soul}
\usepackage{xcolor}
\usepackage{todonotes}
\usepackage{multirow}
\usepackage{xpatch}
\usepackage{blindtext}
\usepackage{fdsymbol}

\usepackage[T1]{fontenc}

\usepackage[utf8]{inputenc}

\usepackage{microtype}
\usepackage{hyperref}
\usepackage{adjustbox}
\usepackage{textcomp}
\usepackage{xparse}
\usepackage{enumitem}

\crefformat{section}{\S#2#1#3} 
\crefformat{subsection}{\S#2#1#3}
\crefformat{subsubsection}{\S#2#1#3}
\definecolor{gold}{rgb}{0.83, 0.69, 0.22}

\iffalse  
\NewDocumentCommand{\heng}
{ mO{} }{\textcolor{red}{\textsuperscript{\textit{Heng}}\textsf{\textbf{\small[#1]}}}}
\NewDocumentCommand{\xueqing}
{ mO{} }{\textcolor{cyan}{\textsuperscript{\textit{Xueqing}}\textsf{\textbf{\small[#1]}}}}
\NewDocumentCommand{\steeve}
{ mO{} }{\textcolor{gold}{\textsuperscript{\textit{Steeve}}\textsf{\textbf{\small[#1]}}}}
\NewDocumentCommand{\yi}
{ mO{} }{\textcolor{green}{\textsuperscript{\textit{Yi}}\textsf{\textbf{\small[#1]}}}}
\NewDocumentCommand{\preslav}{ mO{} }{\textcolor{blue}{\textsuperscript{\textit{Preslav}}\textsf{\textbf{\small[#1]}}}}

\else
\newcommand{\heng}[1]{}
\newcommand{\xueqing}[1]{}
\newcommand{\yi}[1]{}
\newcommand{\steeve}[1]{}
\newcommand{\preslav}[1]{}
\fi

\newlength{\Width}%
\newlength{\DepthReference}
\newlength{\HeightReference}
\newcommand{\MyColorBox}[2][red]%
{%
    \settowidth{\Width}{#2}%
    \colorbox{#1}%
    {%
        \raisebox{-\DepthReference}%
        {%
                \parbox[b][\HeightReference+\DepthReference][c]{\Width}{\centering#2}%
        }%
    }%
}
\DeclareMathOperator*{\argmin}{argmin}
\newcommand{\datasetname}[1]{\textsc{PropaNews}}
\newcommand{\humanwritten}[1]{\textsc{HumanNews}}
\newcommand{\fakeevent}[1]{\textsc{FakeEvent}}
\newcommand{\grover}[1]{\textsc{Grover}}
\newcommand{\grovergen}[1]{\textsc{Grover-gen}}
\newcommand{\factgen}[1]{\textsc{FactGen}}

\definecolor{c2}{RGB}{218,0,0}

\definecolor{lightblue}{RGB}{212, 235, 255}
\definecolor{lightorange}{RGB}{255, 204, 168}
\definecolor{lightyellow}{RGB}{255, 255, 168}
\sethlcolor{lightblue}

\newcommand\hlc[2]{\sethlcolor{#1} \hl{#2}}

\title{Faking Fake News for Real Fake News Detection: \\Propaganda-loaded Training Data Generation}

\author{Kung-Hsiang Huang\textsuperscript{$\spadesuit$}~~~
Kathleen McKeown\textsuperscript{$\clubsuit$}~~~ \\
    {\bfseries 
     Preslav Nakov\textsuperscript{$\diamondsuit$} ~~~
     Yejin Choi\textsuperscript{$\heartsuit\vardiamondsuit$}~~~ 
    Heng Ji\textsuperscript{$\spadesuit$}} \\
  \textsuperscript{$\spadesuit$} UIUC ~~
  \textsuperscript{$\clubsuit$} Columbia University ~~
  \textsuperscript{$\heartsuit$} University of Washington ~~
  \textsuperscript{$\diamondsuit$} MBZUAI ~~
  \textsuperscript{$\vardiamondsuit$}AI2 \\
  \texttt{\{khhuang3, hengji\}@illinois.edu} ~~~ \texttt{kathy@columbia.edu} \\
\texttt{preslav.nakov@mbzuai.ac.ae}~~~ \texttt{yejinc@allenai.org} \\
  }

\begin{document}
\maketitle

\begin{abstract}

Despite recent advances in detecting fake news generated by neural models, their results are not readily applicable to effective detection of human-written disinformation. What limits the successful transfer between them is the sizable gap between machine-generated fake news and human-authored ones, including the notable differences in terms of style and underlying intent. With this in mind, we propose a novel framework for generating training examples that are informed by the known styles and strategies of human-authored propaganda.  
Specifically, we perform self-critical sequence training guided by natural language inference to ensure the validity of the generated articles, while also incorporating propaganda techniques, such as \emph{appeal to authority} and \emph{loaded language}.  In particular, we create a new training dataset, \datasetname~, with 2,256 examples, which we release for future use. Our experimental results show that fake news detectors trained on \datasetname~ are better at detecting human-written disinformation by 3.62--7.69\% F1 score on two public datasets. \footnote{The code and data released on GitHub: \url{https://github.com/khuangaf/FakingFakeNews}}
\end{abstract}
\section{Introduction}
\label{sec:intro}

The dissemination of false information can cause chaos, hatred, and trust issues, and can eventually hinder the development of society as a whole \cite{dewatana2021effectiveness, wasserman2019exploratory}. In particular, human-written disinformation\footnote{There are many types and definitions of \textit{fake news}, but here we focus on text-only \emph{disinformation}. Yet, we will also use the less accurate term \emph{fake news} as it is more common.} is often used to manipulate certain populations and had a catastrophic impact on multiple events, such as Brexit \cite{bastos2019brexit}, the COVID-19 pandemic \cite{van2020inoculating}, and the 2022 Russian assault on Ukraine.

Hence, there is an urgent need for a defense mechanism against human-written disinformation.\footnote{WARNING: This paper contains disinformation that may be sensitive or offensive in nature.} To construct such a mechanism, we need a substantial amount of training data to train the detectors. A na\"{i}ve solution is to collect human-written news articles that contain inaccurate information by crawling untrustworthy news media. However, news articles published by suspicious sources do not necessarily contain false information, which means that annotators are required to fact-check every claim in each untrustworthy article.  Moreover, articles containing false claims are often removed shortly after posting. While some work collected human-written fake news from fact-checking websites \cite{shu2018fakenewsnet,nguyen2020fang}, the size of these datasets is limited. The curation process of these websites also requires high manual efforts. Hence, such a solution is neither scalable nor reliable. Thus, an alternative direction complementing the existing efforts would be generateing training data automatically in a way that avoids these issues.

\begin{table*}[thb]
    \small
    \centering
    {
    \begin{tabularx}{\textwidth}{X}
        \toprule
        AJDABIYAH , Libya | Thu Apr 7 , 2011 6:34 pm EDT AJDABIYAH , Libya -LRB- Reuters -RRB- - Rebels fighting to overthrow Muammar Gaddafi said five of their fighters were killed ... ''In rebel-held eastern Libya, wounded rebels being brought to a hospital Ajdabiyah said their trucks and tanks were hit on Thursday by a NATO air strike outside Brega. \st{NATO said it was investigating an attack by its aircraft on a tank column in the area along the Mediterranean coast on Thursday , saying the situation was `` unclear and fluid . ''} \hlc{lightorange}{Rebels said at least five of their fighters were killed when NATO planes mistakenly bombed a rebel tank column near the contested port.} \hlc{lightblue}{``A number of vehicles were hit by a NATO strike '', officers from UN concluded.} The fighting for Brega , the only active front , has dragged on for a week ...\\

        \bottomrule
    \end{tabularx}
    }
    \vspace{-2mm}
    \caption{An example of our generated fake news. Given an authentic news article, our approach first identifies \st{a salient sentence}, which it then replaces with \hlc{lightorange}{a plausible but disinformative sentence} that is coherent to the context. Finally, it generates a \hlc{lightblue}{propaganda} sentence to make the article resemble human-written fake news. }
    \vspace{-5mm}
    \label{tab:overview_example}
    
\end{table*}

Our goal here is to enhance disinformation detection by generating training examples that are better informed by the known styles and strategies of human-authored disinformation. We started by collecting human-written disinformative articles from untrustworthy sites\footnote{These news sources are rated \textit{low} for the factuality of reporting by \url{mediabiasfactcheck.com}.}, and we analyzed around 40 of them that spread false claims. Throughout our analysis, we found two characteristics of this human-written disinformation. First, about 33\% of the articles used propaganda techniques to convince the audience that the fake information was actually authentic, and these techniques often involve the use of emotion-triggering language or logical fallacies \cite{da-san-martino-etal-2019-fine} to increase the impact on the reader. The count of each propaganda technique used is shown in \Cref{apx:propa_dist}. Second, more than 55\% of the articles that we analyzed contained inaccurate information mixed with the correct information: in fact, all claims, except for one or two, in these disinformation articles were factual, which makes the few false claims in these articles even more believable. 

Prior work has made significant progress in generating fake news using large pre-trained sequence-to-sequence (seq2seq) models \cite{zellers2019grover, fung-etal-2021-infosurgeon, Shu_Li_Ding_Liu_2021}.  However, the articles generated by these approaches contain an overwhelmingly large proportion of false information and do not explicitly use propaganda.

To address these issues, here we propose a novel generation method. Given an authentic news article, we replace a salient sentence with a plausible but fake piece of information using a seq2seq model. As the generated texts can often be entailed by the original contexts, we incorporate a self-critical sequence training objective \cite{rennie-et-al-2017-self} that incorporates a natural language inference (NLI) model into the loss function. 
Additionally, we use the NLI model to filter out generated sentences that can be inferred from the replaced ones. 
Then, we add propaganda techniques to mimic how humans craft disinformation. In particular, 
we automate two commonly used propaganda techniques, \textit{appeal to authority} and \textit{loaded language}, \cite{da-san-martino-etal-2019-fine} to add propaganda into the faked sentences. 

Subsequently, we use the silver-standard training data generated from these two steps to train a detector. An example is shown in \Cref{tab:overview_example}. We further recruited crowdsourcing workers to validate that some of the generated texts were indeed fake, so that we could construct a gold-standard training dataset.

Comparing our method to state-of-the-art fake news generation approaches, the evaluation results on two human-written fake news datasets show that detectors are substantially better at spotting human-written disinformation when trained on our generated fake news dataset. Our ablation studies confirm the effectiveness of incorporating propaganda into the generated articles for producing better training data.

Our contributions can be summarized as follows:
\begin{itemize}[noitemsep,nolistsep]
  \item We propose an effective method to automatically generate more realistic disinformation compared to previous work.
  \item We develop the first automatic methods to generate specific propaganda techniques such that the generated articles are closer to disinformation written by humans.
  \item We demonstrate that detectors trained on our generated data, compared to generated articles using other methods, are better at detecting human-written disinformation.  
  \item We release ~\datasetname~, a dataset for disinformation detection containing 2.2K articles generated by our approach and validated by humans.
\end{itemize}

\section{Training Data Generation}
\label{sec:generation_method}

Our process of generating training data for propaganda-loaded disinformation consists of two main steps: disinformation generation (\Cref{subsec:disinfo_gen}) and propaganda generation (\Cref{subsec:propa_gen}).  
Below, we describe each of these steps in detail.

\subsection{Disinformation Generation}
\label{subsec:disinfo_gen}
Our disinformation generation approach aims at two sub-goals: (\emph{i})~replacing a salient sentence in the given article with a sequence of generated coherent texts that looks plausible, and (\emph{ii})~ensuring that the generated information cannot be entailed by the original masked-out sentence; 
otherwise, the generated texts will not be disinformative. To achieve the first sub-goal, we first identify salient sentences using extractive summarization, and we then perform mask-infilling with \textsc{BART} \cite{lewis-etal-2020-bart}. The second sub-goal is accomplished using self-critical sequence training \cite{rennie-et-al-2017-self} with an NLI component, which is used as a reward function for generation.

\begin{figure*}[bt]
    \centering
    \includegraphics[width=0.9\linewidth]{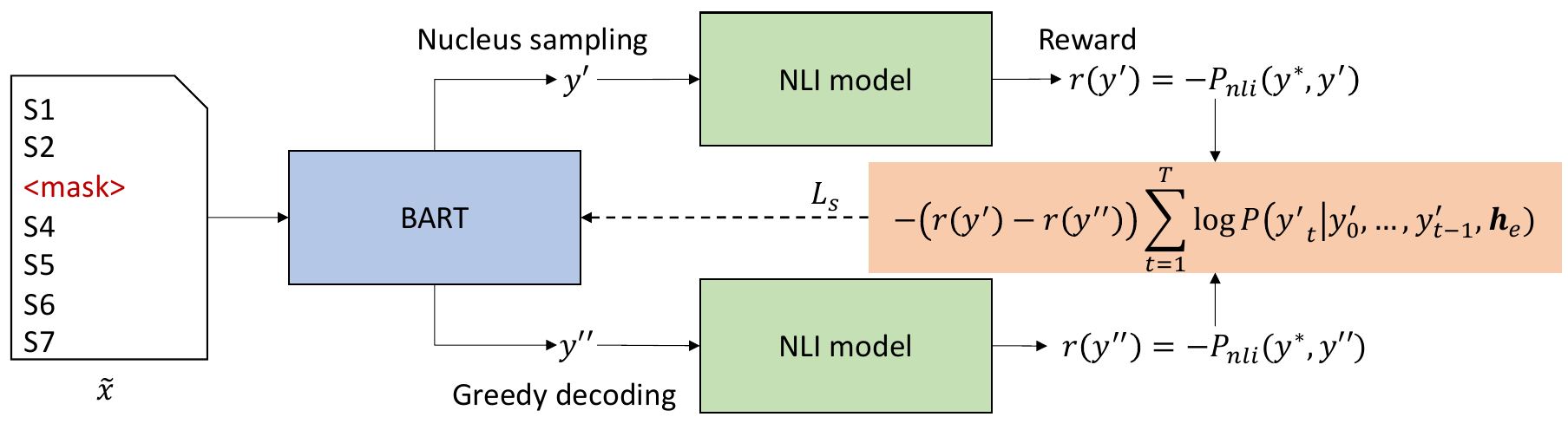}
    \vspace{-2mm}
    \caption{Illustration of our self-critical sequence training. 
    Given a corrupted input article $\Tilde{x}$, \textsc{BART} generates two sequences with Nucleus sampling and greedy decoding, respectively. The reward for each sequence is computed as the negative entailment probability $-P_{ent}$ as output from the NLI model.}
    \vspace{-5mm}
    \label{fig:SCST}
\end{figure*}

\paragraph{Salient Sentence Identification}
A salient sentence is critical for the overall semantics of the article. When a salient sentence is manipulated or replaced, the complex events  described in the article may be drastically changed. Yet, there is no salient sentence identification dataset publicly available. Motivated by the fact that sentences included in an extractive summary are often of higher importance, we take the scores computed by an extractive summarization model \cite{liu-lapata-2019-text}, which predicts how likely each sentence is to belong to the summary, to estimate the saliency of each sentence. Empirically, we found that this approach yields reasonably good sentence saliency estimation. For each news outlet, we replace one sentence that has the highest extractive summarization score with our generated disinformation.

\paragraph{Mask Infilling with \textsc{BART}}
To perform infilling, we take an approach that is similar to that of \citet{donahue-etal-2020-enabling}, but we use \textsc{BART} \cite{lewis-etal-2020-bart}, a pre-trained language model with an encoder--decoder architecture. During training time, we randomly mask out a sentence $y^*$ from a given article $x$. The bidirectional encoder first produces contextualized representations $\bm{h_e} = \mathrm{Encoder}(\Tilde{x})$ given the article with a masked-out sentence $\Tilde{x} = x - y^*$. Then, the auto-regressive decoder learns a maximum likelihood estimation that aims to maximize the probability of generating the next token $y^*_t$ at time step $t$ given all tokens in previous time steps $\{y^*_0, ..., y^*_{t-1 }\}$ 

and the encoder hidden states $\bm{h_e}$ by minimizing the negative log probability of generating $y^*_t$ as follows:
\begin{align}
    \mathcal{L}_{m} = - \sum^{T}_{t=1}\log P(y^*_t| y^*_0, ..., y^*_{t-1}, \bm{h_e}).
    \label{eq:mle_loss}
\end{align}
 
During inference time, rather than random masking, $\Tilde{x}$ is formed by masking out the sentence with the highest score computed by the extractive summarization model given the original document $x$, as discussed in the previous paragraph.

\begin{figure}[t]
    \centering
    \includegraphics[width=0.9\linewidth]{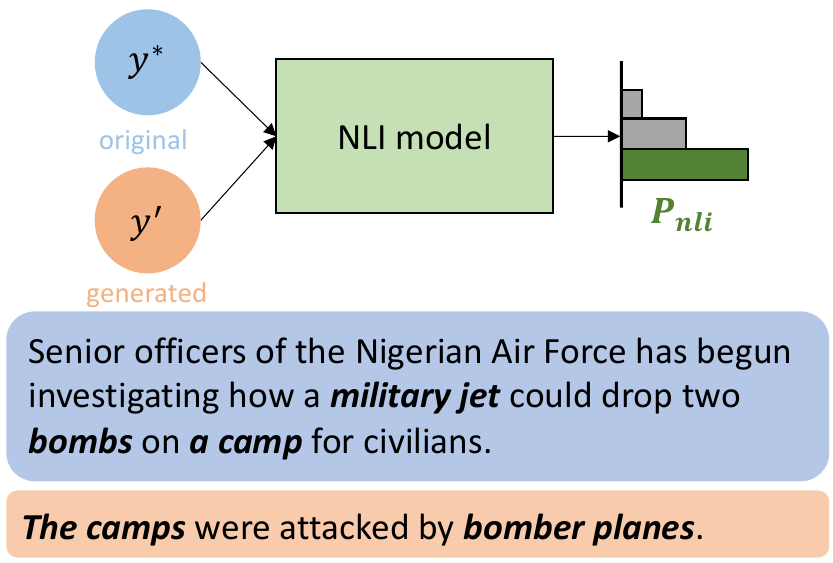}
    \vspace{-2mm}
    \caption{An example showing the NLI model predicts an entailment from the masked out sentence $y^*$ to the generated sentence $y'$. 
    }
    \vspace{-5mm}
    \label{fig:nli_rejected}
\end{figure}

\paragraph{Self-critical Sequence Training}
\textsc{BART} optimized via maximum likelihood estimation alone is capable of generating coherent texts. However, although the generated texts $y'$ may be very different from the originally masked out sentence $y^*$, there is no guarantee that $y'$ contains incorrect information. If the generated texts $y'$ can be entailed by the masked out sentence $y^*$, then $y'$ is actually not disinformative. An example is shown in \Cref{fig:nli_rejected}. Here, except for the lack of details, the generated sentence $y'$ delivers the same message as the masked out sentence $y^*$. To reduce the probability that $y'$ can be entailed by $y^*$, we leverage self-critical sequence training \cite{rennie-et-al-2017-self, bosselut-etal-2018-discourse} that rewards the model for generating sequences that cannot be entailed by the masked-out sentences.
Self-critical sequence training (SCST) is a form of the REINFORCE algorithm \cite{williams1992simple} that allows direct optimization on non-differentiable functions. Using a baseline output $y''$ of the model to normalize the rewards, SCST avoids the challenge of directly estimating the reward signal or estimating normalization \cite{rennie-et-al-2017-self}. Since our goal is to avoid entailment from $y^*$ to $y'$, we define the reward as the negative entailment probability computed by a \textsc{RoBERTa}-based \cite{liu2019roberta} NLI model fine-tuned on Multi-NLI \cite{williams-etal-2018-broad}\footnote{We use the fine-tuned NLI  model from \url{https://huggingface.co/roberta-large-mnli}. Its accuracy is 90.2\% on the dev set of MNLI, which is on par with state-of-the-art methods.},

{
\vspace{-4mm}
\small
\begin{align}
    r(y') = -P_{nli}(y^*, y'),
\end{align}
}
where $r(y')$ is the reward of the sequence sampled from the current policy $y'$, and $P_{nli}(y^*, y')$ is the probability that $y^*$ entails $y'$. To generate $y'$, we use Nucleus Sampling \cite{Holtzman2020The} with $p=0.96$,  as this sampling method has shown advantages in \textit{open-ended} generation \cite{Holtzman2020The, zellers2019grover}. 

We generate the baseline output $y''$ using greedy decoding, then obtain the entailment probabilities between $y'$ and $y''$ from the NLI model. We then compute the self-critical sequence training loss:

{
\vspace{-4mm}
\small
\begin{align}
    \mathcal{L}_{s} = - (r(y') - r(y'')) \sum^{T}_{t=1}\log P(y'_t|y'_0, .., y'_{t-1}, \bm{h_e}).
    \label{eq:sc_loss}
\end{align}
}
\noindent
Here $r(y'')$ is a baseline reward, and $r(y') - r(y'')$ is a normalized reward. This loss function encourages \textsc{BART} to generate $y'$ when $r(y') > r(y'')$, whereas it suppresses the probability of decoding $y'$ when $r(y') < r(y'')$. An overview of SCST is shown in \Cref{fig:SCST}.

The final objective function to minimize is a weighted sum of \Cref{eq:mle_loss} and \Cref{eq:sc_loss},

{
\vspace{-4mm}
\small
\begin{align}
    \mathcal{L}_{final} = \alpha \mathcal{L}_{m} + \beta \mathcal{L}_{s},
     \label{eq:final_loss}
\end{align}
}
where $\alpha$ and $\beta$ are the weights for each loss\footnote{Empirically, we set $\alpha = 1$ and $\beta = 0.01$.}. 

\paragraph{Post-processing}
To further ensure the quality of the disinformation generated, we reuse the NLI model discussed in the previous paragraph to filter out invalid outputs $y'$ that can be entailed from the masked-out sentence $y^*$, as demonstrated in \Cref{fig:nli_rejected}. 
We found that incorporation of the SCST loss (\Cref{eq:sc_loss}) into the training objective successfully reduces the invalid rate from 7.8\% to  3.2\%.

\begin{table*}[thb]
    \small
    \centering
    {
    \begin{tabularx}{\textwidth}{lX}
        \toprule
        \textbf{Technique} & \textbf{Generated Disinformation and Propaganda} \\
        \midrule
        
        \textbf{Appeal to Authority} & Cairo's Tahrir Square was the scene of clashes between protesters and police on Wednesday.{\hlc{lightblue}{`` At least three people were killed and more than 600 were injured in the clashes,'' said Egypt's President.}} \\
        \midrule
        \textbf{Loaded Language} & Cairo's Tahrir Square was the scene of {\hlc{lightblue}{deadly}} clashes between protesters and police on Wednesday. \\
        
        \bottomrule
    \end{tabularx}
    }
    \vspace{-2mm}
    \caption{Examples of the two generated propaganda techniques, as shown by \hlc{lightblue}{texts in blue}. The first row shows how the argument is strengthened by appealing to an authority's statement, while the second row demonstrates how loaded language is introduced with an emotion-triggering term. }
    \vspace{-5mm}
    \label{tab:propaganda_outputs}
    
\end{table*}

\subsection{Propaganda Generation}
\label{subsec:propa_gen}
After generating inaccurate information, we then incorporate propaganda into each generated article. We chose two representative propaganda techniques of each type: emotional versus non-emotional. \textit{Loaded language} is an emotional technique and it is also by far the most frequent propaganda technique as shown in Table 5 of \citet{da-san-martino-etal-2019-fine} and Table 2 of \citet{dimitrov-etal-2021-detecting}. Based on these two tables, we also see that \textit{appeal to authority} is among the most frequent non-emotional techniques.

\paragraph{Appeal to Authority} \textit{Appeal to authority} is a propaganda technique that aims to strengthen or invalidate an argument by referring to a statement made by authorities or experts \cite{da-san-martino-etal-2019-fine}. We first collect experts from various domains, such as economics and immunology, from Wikidata\footnote{\url{https://query.wikidata.org/}}. In particular, we specify the \textit{occupation} (P108) of each expert and filter out entities that were born before 1940 to ensure recency. To consider only impactful entities, we rank all candidates based on the number of corresponding outcoming \textit{statements} (i.e. connected concepts in Wikidata), inspired by PageRank \cite{ilprints422}, and add the top 100 entities for each occupation into the candidate list $Z$. Then, we include the person named entities extracted by a name tagger\footnote{\url{https://stanfordnlp.github.io/stanza}}, which are more relevant to the local context. 

This makes sense as we found that more than 73\% of the news articles contain authorities. More details on how authority candidates $Z$ are collected can be found in \Cref{apx:app_auth_details}.

Once we collect a candidate list $Z$, we then generate fake arguments made by each $z_i \in Z$  with the \textsc{BART} model that has already been fine-tuned in \Cref{subsec:disinfo_gen}. In particular, a {\tt <mask>} token is inserted right after the filled-in sentence $y'$ in the input article to \textsc{BART} so that it knows where to perform infilling. To inform \textsc{BART} that it should generate a statement made by an  authority, we prefix the decoder with a template such as [$z_i$ confirmed that ``], where $z_i \in Z$ is the name of the authority. 

The prefix ends with an opening quotation mark to indicate that it should be followed by a statement by authority $z_i$. To increase the diversity of the generated statements, we devise a variety of templates, as detailed in \Cref{apx:app_auth_details}. Finally, the best sequence $s^*$ is selected with the lowest perplexity $s^* = \argmin_{s_i} \textrm{Perplexity}(s_i) $, where $s_i$ denotes the generated sequence using $z_i$ as the authority.

\paragraph{Loaded Language} \textit{Loaded language} is another propaganda technique that uses emotion-triggering terms or phrases to influence the opinions of the audience \cite{da-san-martino-etal-2019-fine, dimitrov-etal-2021-detecting}. Often, \textit{loaded language} involves the use of sensational adverbs or adjectives to exaggerate a statement. Based on this observation, we utilize the propaganda dataset released by \citet{da-san-martino-etal-2019-fine} where propaganda techniques are annotated at the fragment level (i.e. span level). The dataset contains 2,547 \textit{loaded language} instances. Yet, not every instance contains adjectives or adverbs that are emotion-triggering. To create valid training data for \textit{loaded language} generation, we first use SpaCy to perform part of speech tagging and dependency parsing, and then keep the examples where there exists an adverb pointing to a verb or an adjective pointing to a noun through dependency parsing edges. This results in 1,017 samples of valid \textit{loaded language} instances. Examples of the generated \textit{appeal to authority} and \textit{loaded language} are shown in \Cref{tab:propaganda_outputs}. 

Upon collecting the training data to generate \textit{loaded language}, we fine-tune another \textsc{BART} on this dataset. Na\"{i}vely, we can take the articles with emotion-triggering adverbs or adjectives removed as input to \textsc{BART} and using the original article as the decoding target. However, we found that around 25\% of the time \textsc{BART} does not exactly reproduce
the unmasked texts due to hallucination. This observation is consistent with \citet{donahue-etal-2020-enabling}'s findings. To this end, we propose a two-step generation approach. First, we train \textsc{BART} to insert a {\tt <mask>} token into the target sentence in the input document marked with special tokens. Then, \textsc{BART} learns to infill the {\tt <mask>} with an approach similar to what is discussed in \Cref{subsec:disinfo_gen} but without the SCST objective. Empirically, we found that this approach successfully reduces the chance of failure in generating the exact unmasked contexts to around 2\%. 

\subsection{Intermediate Pre-training} 
As the size of \textsc{Timeline17} \cite{Tran2013LeveragingLT} and the propaganda dataset \cite{da-san-martino-etal-2019-fine} are relatively small, we perform intermediate pre-training (IPT) on the news articles from \textsc{CNN/DM}, a large news summarization dataset \cite{Hermann2015TeachingMT}, for domain adaptation. Details of IPT can be found in \Cref{apx:ipt}.


\section{Our \datasetname~ Dataset}

\subsection{Data Source}
When selecting the source of data to construct our dataset, we consider the following two criteria. First, the news articles must have high trustworthiness. This ensures that, except for our manipulated sentences, the rest of the articles are genuine. Second, the news events described in the articles must be important to the general audience. Motivated by these two criteria, we repurpose the \textsc{Timeline17} dataset \cite{Tran2013LeveragingLT} as our source of data. \textsc{Timeline17} contains 17 timelines, each of which corresponds to a news event. Each timeline is associated with a series of news articles that span across a wide time span, implying the high importance and impact of these news events. Additionally, the news articles are from trustworthy media, such as The New York Times and The Guardian. In total, there are 4,535 news articles in \textsc{Timeline17}.

\subsection{Crowdsourcing for Data Curation} 
\label{subsec:crowdsourcing}
We use Amazon's Mechanical Turk (AMT) to verify the quality and correctness of the generated disinformation. In total, there are around 400 unique crowdsourcing workers contributing to approximately 2,000 Human Intelligence Tasks (HITs). For each HIT, annotators are tasked to look for supporting evidence from trustworthy news media to determine whether the sentences generated are indeed \textit{inaccurate}. Only those labeled as \textit{inaccurate} will be included in \datasetname~, while the \textit{accurate} counterparts are discarded. 
 \Cref{apx:annotation_interface} provides the details of the annotation interface.

To measure the inter-annotator agreement (IAA), we use the Worker Agreement With Aggregate (WAWA) score, following \citet{ning-etal-2020-torque} and \citet{sheng-etal-2021-nice}. WAWA compares each annotator's answer with the aggregated answer obtained via majority votes and micro-averages the results across all samples\footnote{We did not use other IAA metrics, such as Cohen's Kappa \cite{cohen1960coefficient}, as we expect the vast majority of our generated disinformation to be inaccurate. WAWA provides a better approximation for inter-annotator agreement in our scenario.}. The resulting WAWA precision, recall, and $\mathrm{F}_1$ are 80.01\%, 78.94\%, and 79.47\%, which indicates a moderate to high agreement. \looseness=-1
\section{Disinformation Detection}
The disinformation detection task challenges detectors to determine whether a given input article contains inaccurate information or not. We experiment on four detectors, including \textsc{HDSF} \cite{karimi-tang-2019-learning}, \grover~ \cite{zellers2019grover}, \textsc{Bert} \cite{devlin-etal-2019-bert} and \textsc{RoBERTa} \cite{liu2019roberta}. \textsc{HDSF} leverages the hierarchical structures of discourse-level features, such as dependency trees, to predict the veracity of a news article. \grover~ is an  unidirectional seq2seq model pre-trained on news  documents. We use the discriminative version for detection which is adapted from its generative version by feeding the \texttt{[CLS]} token representations to a multi-layer perceptron. Similarly, \textsc{Bert} and \textsc{RoBERTa} take in the entire article as input and feed the representations of the first token to a classification head to determine the veracity of each article. In addition, all models are optimized using cross entropy. For fair comparison, we set the maximum sequence length to 512 and use the \textsc{Large} variants for all models. Details can be found in \Cref{apx:implementation_details}. 
\section{Experiments}
\heng{I suggest to move some part of table 8 to main text if all possible. the current experiments section without examples is a bit dry. Use a different name for -1Sent?}

In our experiments, we aim to (1) analyze the performance of different models on the \datasetname~ dataset, (2) examine the effect of various training data sets, and (3) investigate how much silver-standard data is equivalent to gold-standard data.
\begin{table*}[t]
    \small
    \centering
    \begin{adjustbox}{max width=\linewidth}
    {
    \begin{tabular}{lcccc}
        
        \toprule
        \textbf{Test Data $\rightarrow~$} & \multicolumn{2}{c}{\textbf{\textsc{PolitiFact}}} & \multicolumn{2}{c}{\textbf{\textsc{Snopes}}}\\
        \cmidrule(lr){2-3} \cmidrule(lr){4-5}
        \textbf{Detectors $\rightarrow~$} & \textsc{RoBERTa-Large}  & \textsc{Grover-Large} & \textsc{RoBERTa-Large}  & \textsc{Grover-Large} \\
        
        \textbf{Training Data $\downarrow~$}      \\
        \midrule
        \multicolumn{5}{c}{Without human validation (silver)}\\
        
        \midrule

        \grovergen~  & \multicolumn{1}{c}{57.65} ($\pm$7.6) & \multicolumn{1}{c}{52.77} ($\pm$2.1) & \multicolumn{1}{c}{48.42} ($\pm$2.2) & \multicolumn{1}{c}{49.53} ($\pm$0.1) \\
        $\grovergen~{\textsc{-1Sent}}$ & \multicolumn{1}{c}{49.65} ($\pm$5.2) & \multicolumn{1}{c}{47.48} ($\pm$1.8) & \multicolumn{1}{c}{44.44} ($\pm$3.2) & \multicolumn{1}{c}{50.10} ($\pm$2.1) \\
        \fakeevent~ & \multicolumn{1}{c}{46.33} ($\pm$2.6) & \multicolumn{1}{c}{50.27} ($\pm$5.9) & \multicolumn{1}{c}{45.36} ($\pm$1.2) & \multicolumn{1}{c}{47.40} ($\pm$1.3) \\
        $\fakeevent~{\textsc{-1Sent}}$ & \multicolumn{1}{c}{47.32} ($\pm$3.2) & \multicolumn{1}{c}{50.12} ($\pm$3.2) & \multicolumn{1}{c}{46.62} ($\pm$2.9) & \multicolumn{1}{c}{47.29} ($\pm$2.7) \\
        \factgen~ & \multicolumn{1}{c}{48.46} ($\pm$2.2) & \multicolumn{1}{c}{51.79} ($\pm$3.6) & \multicolumn{1}{c}{41.98} ($\pm$5.4) & \multicolumn{1}{c}{50.47} ($\pm$4.9) \\ 
        $\factgen~\textsc{-1Sent}$ & \multicolumn{1}{c}{41.19} ($\pm$3.5) & \multicolumn{1}{c}{40.92} ($\pm$4.1) & \multicolumn{1}{c}{40.01} ($\pm$3.8) & \multicolumn{1}{c}{45.52} ($\pm$3.7) \\
        \textsc{PN-silver}~ & \multicolumn{1}{c}{~~\textbf{60.39}$^{*}$} ($\pm$3.9) & \multicolumn{1}{c}{~~\textbf{55.23}$^{*}$} ($\pm$5.8) & \multicolumn{1}{c}{~~~~\textbf{51.52}$^{**}$} ($\pm$3.4) & \multicolumn{1}{c}{~~~~\textbf{52.39}$^{**}$} ($\pm$4.1) \\
        
        \midrule
        \multicolumn{5}{c}{With human validation (gold)}\\
        \midrule
        \datasetname~ & \multicolumn{1}{c}{~~~~\textbf{65.34}$^{**}$} ($\pm$4.5) & \multicolumn{1}{c}{~~~~\textbf{60.43}$^{**}$} ($\pm$6.2) & \multicolumn{1}{c}{~~~~\textbf{53.03}$^{**}$} ($\pm$3.7) & \multicolumn{1}{c}{~~~~\textbf{54.09}$^{**}$} ($\pm$2.8) \\
        ~~ w/o AA & \multicolumn{1}{c}{~~~~63.21$^{**}$} ($\pm$3.2) & \multicolumn{1}{c}{~~~~58.28$^{**}$} ($\pm$4.2) & \multicolumn{1}{c}{~~50.78$^{*}$} ($\pm$1.8) & \multicolumn{1}{c}{~~~~53.22$^{**}$} ($\pm$3.7) \\
        ~~ w/o LL & \multicolumn{1}{c}{~~~~64.65$^{**}$} ($\pm$1.8) & \multicolumn{1}{c}{~~~~56.93$^{**}$} ($\pm$5.3) & \multicolumn{1}{c}{~~~~51.92$^{**}$} ($\pm$3.4) & \multicolumn{1}{c}{~~51.68$^{*}$} ($\pm$1.4) \\
        ~~ w/o AA \& LL & \multicolumn{1}{c}{~~61.83$^{*}$} ($\pm$4.9) & \multicolumn{1}{c}{52.82} ($\pm$3.3) & \multicolumn{1}{c}{~~~~52.77$^{**}$} ($\pm$2.7) & \multicolumn{1}{c}{50.93} ($\pm$2.7) \\

        \bottomrule
    \end{tabular}
    }
    \end{adjustbox}
    \vspace{-2mm}
    \caption{AUC (in \%) of different models on the \textsc{Snopes} and \textsc{PolitiFact} datasets when trained on various data sets. The bottom rows show different variants of \datasetname~. AA denotes \textit{appeal to authority}, whereas LL refers to \textit{loaded language}.  We report the mean and standard deviation of four runs. Statistical significance over previous best approaches computed using the paired bootstrap procedure \cite{berg-kirkpatrick-etal-2012-empirical} is indicated with $^{**}$($p < .01$) and $^{*}$($p < .05$) .}
    \vspace{-7mm}
    \label{tab:human_written_test_results}
    \
\end{table*}
\subsection{Data}
\paragraph{\datasetname~} The \datasetname~ dataset consists of 2,256 distinct articles, with a balanced portion of fake and real documents. Within the fake articles, 30\% of them use \textit{appeal to authority}, another 30\% include \textit{loaded language}, and the remaining 40\% simply contains inaccurate information.  We split the data into 1,256: 500: 500 for training, validation, and testing.

\paragraph{Evaluation Data} We use two sets of human-written articles released by \citet{nguyen2020fang} and \citet{shu2018fakenewsnet} to evaluate the effectiveness of our approach. The articles in each dataset are collected from two fact-checking websites, \textsc{Snopes} and \textsc{PolitiFact}, respectively. Articles no longer accessible via the given URL are removed. The statistics of both datasets are shown in \Cref{apx:dataset_stats}.

\paragraph{Other generated training data} We compare \datasetname~ with the following approaches. \textbf{\grovergen~} \cite{zellers2019grover} generates headlines which condition on the original body texts, followed by body text generation conditioning on the generated headlines. \textbf{\factgen~}  \cite{Shu_Li_Ding_Liu_2021} enhances the factual consistency of the generated article with a fact retriever that fetches supporting information from external corpora. \textbf{\fakeevent~} \cite{wu-etal-2022-cross} generates sentences sequentially with condition on the manipulated knowledge elements of each sentence. Also, we form the \textbf{\textsc{PN-silver}} dataset by resampling our generated data but disregarding the annotator validation. Furthermore, we construct additional training sets by replacing the salient sentence in each article with one sentence generated by each baseline method, as indicated by \textbf{\textsc{-1Sent}}. To ensure fair comparisons, all generators take in the same set of authentic articles as inputs.

\subsection{Results and Discussion}
\label{subsec:results}

\paragraph{Human-written disinformation detection} To study the effectiveness of human-written disinformation detection, we train \textsc{Grover-Large} and \textsc{RoBERTa-Large} on different training datasets and evaluate them on the \textsc{Snopes} and \textsc{PolitiFact} datasets, as shown in \Cref{tab:human_written_test_results}. Both models perform best when trained on \datasetname~, compared to training on other datasets. Consider ablating human validation, detectors trained on \textsc{PN-silver} still outperform their counterparts trained on other datasets. This shows that our generative method produces articles that are more similar to human-written disinformation. To further verify this finding, we measure the similarity between articles generated by different approaches and disinformative articles in the \textsc{PolotiFact} dataset using the \textsc{Mauve} metric \cite{pillutla-etal:mauve:neurips2021}. \textsc{Mauve} computes the similarity between two text distributions by adding the areas under a divergence curve, and has been shown to produce better approximations than other metrics such as JS divergence \cite{martins-etal-2020-sparse}.
We find that the \textsc{Mauve} score with \textsc{PolotiFact} for \datasetname~ and \textsc{Grover-gen} are 17.1\% and 13.7\%, respectively, suggesting that the generated documents in \datasetname~ are closer to human-written disinformation. These results confirm that the advantage of our generated articles in defending against human-written disinformation is resulted from the closer gap between them. 

Comparing each baseline method and its counterpart that only generates one sentence to be substituted for the salient sentence (i.e. \textbf{\textsc{-1Sent}}), we found significant performance drops on \grovergen~ and \factgen~ when only generating one sentence. This is likely caused by the incoherence between the right context and the sentence generated by these approaches due to the left-to-right fashion of text generation. While \fakeevent~ does not see the right context, it additionally conditions on knowledge elements corresponding to the sentence, which discourages it from producing topically irrelevant content and thus does not lead to huge performance drop.

In \Cref{tab:qualitative_analysis}, we show two disinformative articles from \textsc{PolitiFact} where \textsc{RoBERTa} is able to classify them as inaccurate when trained on \textsc{PN-silver} but fails when trained on \textsc{Grover-gen}. Both articles contain propaganda, which are incorporated into  \textsc{PN-silver} but not into \textsc{Grover-gen}. This demonstrates that detectors trained on our generated data are better at detecting human-written disinformation that has such properties.

\paragraph{Is propaganda generation helpful for disinformation detection?}

We further conduct an ablation study to analyze the contributions of each propaganda technique. As shown in the bottom of \Cref{tab:human_written_test_results}, both \textit{appeal to authority} and \textit{loaded language} prove beneficial in enhancing models' abilities to detect human-written disinformation. Furthermore, comparing \textsc{PropaNews w/o AA\& LL} with other generation approaches, we find that both models trained on our generated data, even without the incorporation of propaganda techniques, still outperform their counterparts trained on other datasets. This illustrates that our generated disinformation is closer to those written by humans.
\begin{table*}[t]
    \setlength{\tabcolsep}{2pt}
    \small
    \centering
    
    {
    \begin{tabular}{p{0.95\linewidth}}
        \toprule
        \textbf{Article and Analysis} \\
        \midrule

        \textbf{Article:} ... \hl{Statement from FDA Commissioner Scott Gottlieb}, M.D., on FDA’s ongoing efforts to help improve effectiveness of influenza vaccinesFor Immediate Release: ...\\
        
        \textbf{Analysis:}  \textit{Appealing to authority} is common in human-written fake news.\\
        \midrule

        \textbf{Article:} ... Regardless of how much we hate Nacy Pelosi, she represents a Congressional District that saw a million \hl{fraudulent} votes from illegal immigrants... \\
        \textbf{Analysis:}  The use of \textit{loaded language} often indicates disinformation.\\
        \bottomrule
    \end{tabular}
    }
    \vspace{-2mm}
    \caption{Examples from \textsc{PolitiFact} where \textsc{RoBERTa-Large} successfully predicts the veracity when trained on \textsc{PN-silver}, but classifies incorrectly when trained on \textsc{Grover-gen}.  }
    \vspace{-6mm}
    \label{tab:qualitative_analysis}
    
\end{table*}

\paragraph{How good is the generation quality?} To evaluate the quality of our generation approach, we asked AMT workers to rate the plausibility of 100 generated articles from \datasetname~ and determine the degree by which their answer to this question is influenced by the generated propaganda. Each article is rated by 3 workers. For comparison, we also ask AMT workers to rate the plausibility of 100 generated articles from \textsc{Grover-gen}. The average plausibility scores for \datasetname~ and \textsc{Grover-gen}  are 2.25 and 2.15 (out of 3), indicating that our generation approach has a slight advantage over \textsc{Grover-gen} in terms of plausibility. Furthermore, among the articles in \datasetname~ that are rated highly plausible, 29.2\% of the workers think that the generated propaganda highly affects their response (i.e. rated 3 out of 3) that the generated article is plausible. This demonstrates the effectiveness of our propaganda techniques in increasing the plausibility of generated articles. Survey details and score distributions are discussed in \Cref{apx:human_eval_details}.

\section{Related Work}

\paragraph{Fake News Generation and Detection}
 
 There has been a focus in prior research on utilizing neural networks to automatically generate fake news as a means of defending against the proliferation of machine-generated fake news. \citet{zellers2019grover} pre-train a generator with the same architecture as GPT-2 \cite{radford2019language} on a large-scale news corpus and demonstrate that this generator is effective in detecting neural fake news. More recently, \citet{fung-etal-2021-infosurgeon} improve the controllability of the generated fake news by conditioning the generator on knowledge elements, such as entities, relations and events, extracted from the original news article. \citet{Shu_Li_Ding_Liu_2021} enhance the factuality of the generated article by introducing a fact retriever that fetches relevant information from external corpora. \citet{mosallanezhad2021generating} utilize adversarial reinforcement learning to generate topic-preserving articles. These studies have developed methods for generating fake news that is hard to distinguish from real news to humans. 
 Nevertheless, due to the overwhelming amount of inaccurate information introduced and the lack of propaganda techniques in the generated texts, these approaches are sub-optimal for detecting human-written fake news, as shown in \Cref{subsec:results}. In contrast, our work generates fake news by incorporating propaganda techniques and preserving the majority of the correct information. Hence, our approach is more suitable for studying defense against human-written fake news. Also, since our released dataset is annotated with the exact offset of the disinformative passages, this work opens up future research opportunities on interpretable detection of fake news.

\paragraph{Propaganda Generation and Detection}
There is little previous study on propaganda generation. \citet{zellers2019grover} is the only relevant work that we know of that studies the generation of propaganda to communicate targeted disinformation. Our work focuses on generating specific propaganda techniques to bring the generated articles closer to human-written fake news. To the best of our knowledge, we are the first to study the incorporation of specific propaganda techniques into generated articles. 
Prior work on propaganda detection mainly focuses on document-level detection. Early work collects propaganda datasets using distant supervision \cite{rashkin-etal-2017-truth} by assigning the same propaganda label to each news outlet under the same source based on the news-media-level label of corresponding news source listed on trustworthy sites. 
However, classifiers trained on such datasets may only learn to recognize the bias of each news source instead of propaganda \cite{martino2020survey}. Our dataset avoids such issues by explicitly incorporating propaganda into each generated article. Furthermore, \citet{da-san-martino-etal-2019-fine} present a fragment-level propaganda detection dataset, where specific propaganda techniques are labeled onto spans of text instead of each document. Recent approaches for detecting these propaganda techniques rely on pre-trained transformers \cite{morishita-etal-2020-hitachi, feng-etal-2021-alpha}. By contrast, we focus on detecting disinformative articles with propaganda signals.

\section{Conclusions and Future Work}

We have proposed a novel method for generating disinformation that is closer to human-written fake news. Evaluation on two human-written fake news datasets, \textsc{PolitiFact} and \textsc{Snopes}, demonstrates the effectiveness of our generated data \datasetname~ in enabling better detection performance on human-written fake news. We hope that the dataset presented in this work, \datasetname~, can serve as enabling resources for the detection of human-written fake news and encourage future research in this direction. For future work, we plan to extend our approach to other languages and cover more propaganda techniques. We are also interested in studying other aspects of fake news generation, such as novelty and elaboration, as well as engaging linguistic style.
\section{Limitations}
\begin{figure}[b]
    \centering
    \includegraphics[width=0.9\linewidth]{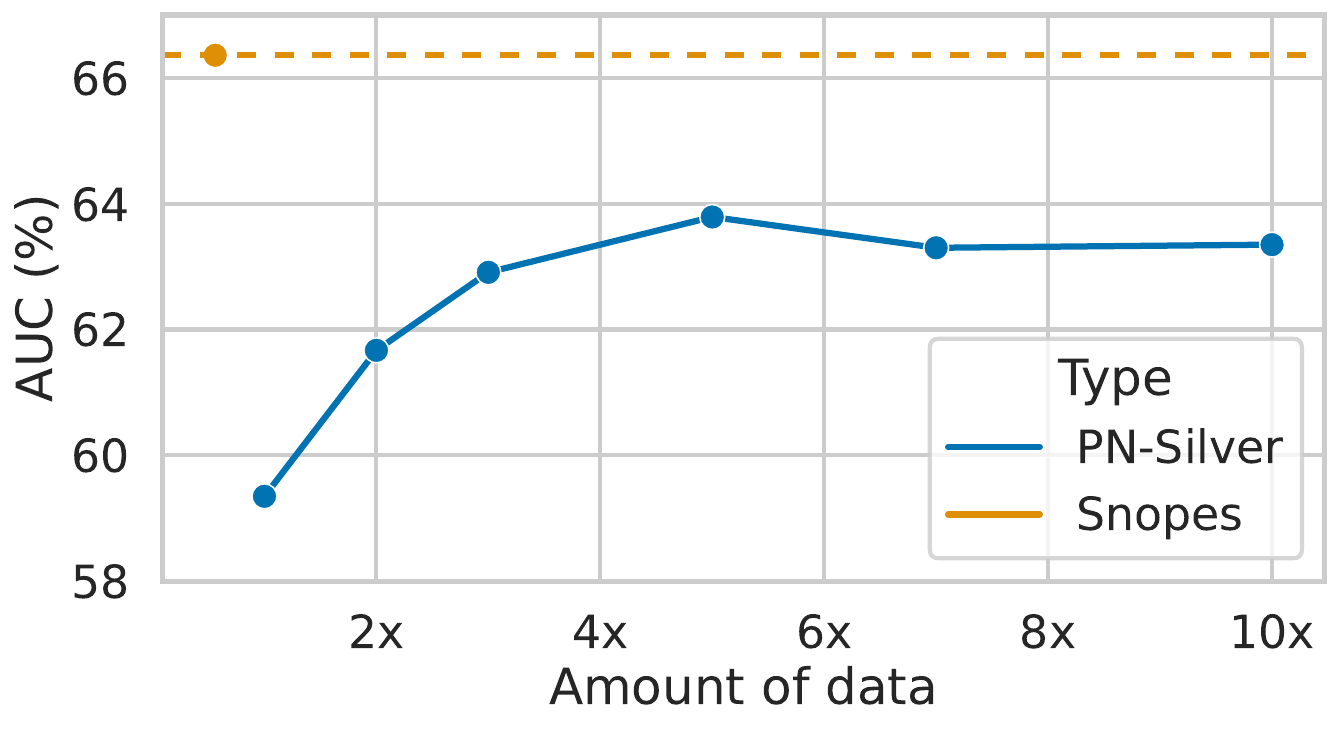}
    \vspace{-2mm}
    \caption{Performance comparison of \textsc{RoBERTa-Large} on the \textsc{PolitiFact} dataset when trained on \textsc{Snopes} and different size of \textsc{PN-silver}.}
    \vspace{-5mm}
    \label{fig:gold2silver}
\end{figure}

To understand the gap between our automatic data generation method and fake news written by humans, we expanded \textsc{PN-silver} to different sizes and compared the performance of \textsc{RoBERTa-Large} between trained on these generated data and the human-written fake news dataset, \textsc{Snopes}. Note that since the \textsc{Timeline17} dataset only contains around 4K samples, we additionally crawled New York Times news articles as input to our generator for the ``5 times'' to ``10 times'' experiments.
The results are shown in \Cref{fig:gold2silver}. Although the detector performance improves as we add more \textit{silver} training data at first, it reaches a plateau after the size is increased to 5 times. This illustrates that while our approach is more effective compared to baseline generation methods, there is still a clear gap between our generated articles and human-crafted fake news, likely in the aspects of styles (as discussed in \Cref{subsec:results}), intents (i.e. limited modeling of propaganda techniques), and falsehood (i.e. the generated content is 100\% false).

Despite the advantages of our generation approach, as compared to previous methods, it is uncapable of generating other propaganda techniques covered in \citep{da-san-martino-etal-2019-fine}, such as \textit{straw man}. Thus, our method is not generic enough to handle all types of propaganda techniques within a unified framework. Moreover, our approach is limited to generating English-only news articles, and cannot be applied to other languages.

\section{Ethical Statement and Broader Impact}

Our objective for developing a generative approach that produces more realistic news articles is to advance the field of disinformation detection and to bring awareness that the current approaches for generating training data for fake news detection are sub-optimal. 

We acknowledge that our generator may produce toxic text as it was fine-tuned on a propaganda datasets. We also understand the dual-use concerns for such a generation framework. One potential concern is the possibility of using the generator to produce fake news for political gain or to sow social discord. Another concern is the potential for the generator to be used to generate fake news that could cause harm, such as false medical information or misleading financial advice. Additionally, the generator might be used to create false evidence or fabricate information to support false allegations in legal or regulatory proceedings. 

Therefore, to contribute to future studies on human-written disinformation detection, we decided to release the codebase for only the detectors used in the experiments as well as the generated data but not the generator.

We highlight some scenarios that illustrate appropriate and inappropriate uses of our generator:
\begin{itemize}[noitemsep,nolistsep,leftmargin=0pt]
\item \textbf{Appropriate:} Researchers can use our framework to produce more challenging training data for learning stronger detectors.
\item \textbf{Inappropriate:} The method should not be used to intentionally create or propagate false information. 
\item \textbf{Inappropriate:} The propaganda generation technique should not be used for political campaigns or any malicious purposes.
\end{itemize}
Both inappropriate uses could lead to harmful consequences, such as undermining trust in the media and causing social unrest.



\section*{Acknowledgement}
This research is based upon work supported by U.S. DARPA SemaFor Program No. HR001120C0123 and DARPA MIPs Program No. HR00112290105. The views and conclusions contained herein are those of the authors and should not be interpreted as necessarily representing the official policies, either expressed or implied, of DARPA, or the U.S. Government. The U.S. Government is authorized to reproduce and distribute reprints for governmental purposes notwithstanding any copyright annotation therein. \looseness=-1

\bibliography{anthology,custom}
\bibliographystyle{acl_natbib}

\clearpage
\appendix

\label{sec:appendix}

\section{Distribution of Propaganda}
\label{apx:propa_dist}
\begin{figure}[h]
    \centering
    \includegraphics[width=0.9\linewidth]{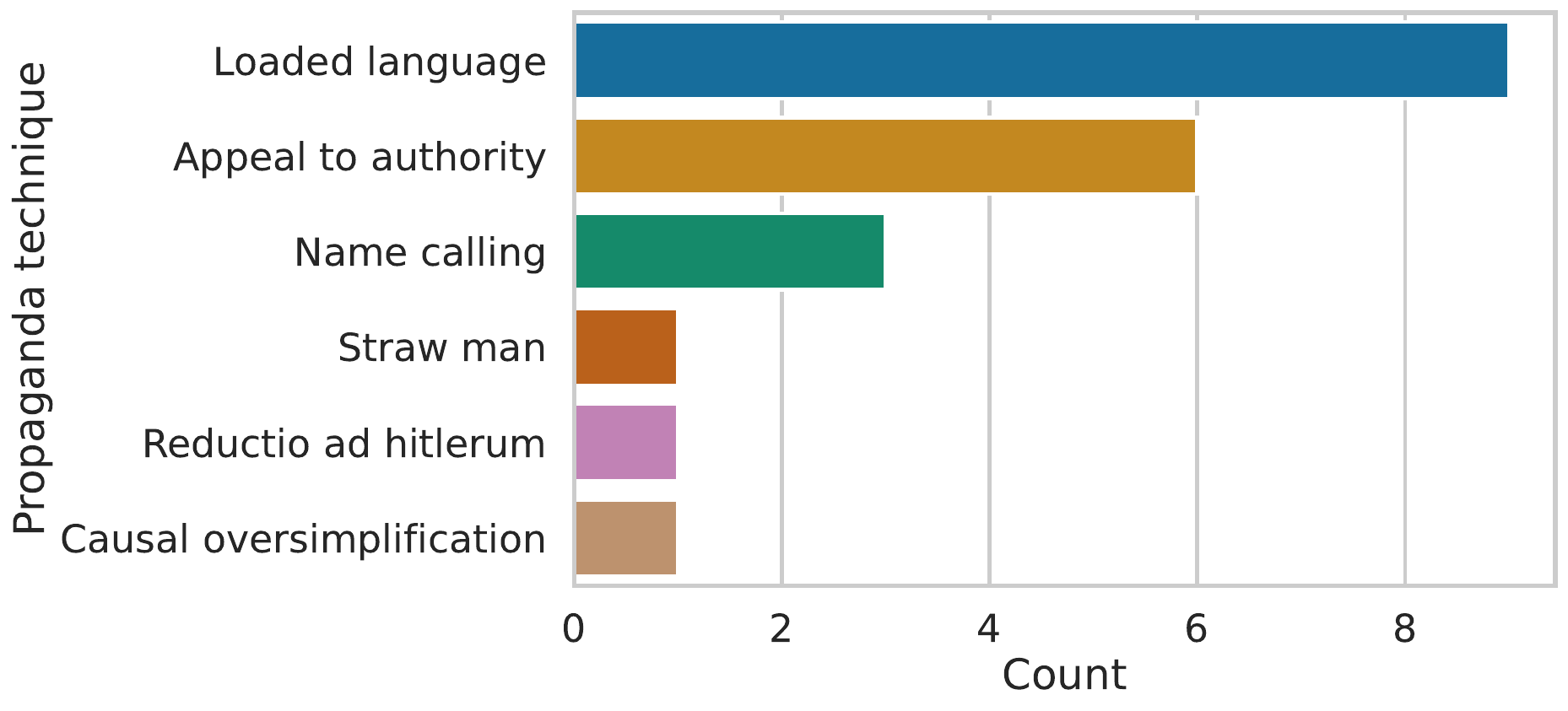}
    \caption{Total counts of the propaganda techniques used in the human-written fake news we analyzed. }
    \label{fig:propaganda_analyzed}
\end{figure}

\Cref{fig:propaganda_analyzed} shows the distribution of the propaganda techniques used in the human-written fake news we collected and analyzed in \Cref{sec:intro}. Note that one article may contain multiple propaganda techniques.
\section{Additional Research Questions}
\paragraph{Q1: Is the detector learning to distinguish between fake/real news articles or simply learning to detect the use of propaganda information?}
In \Cref{tab:human_written_test_results}, \textbf{\datasetname~ w/o AA \& LL} is the variant of our proposed dataset with both propaganda techniques removed. By training detectors on this version of the proposed dataset, the model is still effective in identifying human-written articles containing false information. Therefore, the detectors trained on our generated data have learned to distinguish between fake and real articles instead of exploiting propaganda information only. On the other hand, comparing the detectors trained on \textbf{\datasetname~} and their counterparts trained on \textbf{\datasetname~ w/o AA \& LL} in \Cref{tab:human_written_test_results}, we see that propaganda information can serve as additional evidence that helps improve the detection of real huma-written fake news.

Additionally, we wanted to emphasize that fake news detection is an extremely challenging task that requires both factual and stylistic analysis as demonstrated by our experiments as well as by the relatively low performance of the prior SOTA model.

\paragraph{Q2: Do real articles make use of propaganda techniques, such as \textit{appeal to authority} and \textit{loaded language}?}
The similarity between our generated text and the real articles in PolitiFact is 7.3\% as per the MAUVE metric, which is much lower than the similarity between the generated text and the fake
news articles, as discussed in \Cref{subsec:results}. It is possible that some real news articles can also contain propaganda. However, according to the \textsc{Mauve} metric, the real articles in \textsc{PolitiFact} do not contain much loaded language or appeal to authority.

\section{Further Analysis}
\subsection{Remaining Challenges}

To better understand the remaining disinformative articles that the detectors failed to identify, we conduct an analysis by comparing the \textsc{RoBERTa} predictions and the labels. Three major modeling capabilities required for successful detection are identified, as listed below:

\paragraph{Static knowledge enrichment} About 30\% of the misclassification is resulted from the lack of static knowledge that can be found in public databases, such as law dictionaries. For example, 
in this article\footnote{\url{https://tinyurl.com/static-knowledge}}, Alexandria Ocasio Cortez falsely states that the U.S. Immigration Customs Enforcement (ICE) is required to fill 34,000 beds every day. According to the Appropriations Act of 2016\footnote{\url{https://www.congress.gov/114/bills/hr2029/BILLS-114hr2029enr.pdf}}, however, ICE is only required to detain 34,000 available beds. Therefore, to detect such a misinformation, the detector needs to be enriched with static knowledge bases.

\paragraph{Dynamic knowledge acquisition} Around 48\% of the misclassified human-written disinformation are caused by the inability to acquire dynamic knowledge from new news sources. For instance, COVID-related articles are usually published after 2020, while \textsc{RoBERTA} was pre-trained on news articles released before 2019. It is very challenging for \textsc{RoBERTA} to detect disinformation of such topics unless the detector is equipped with the capabilities of acquiring dynamic knowledge from news articles. Particularly, \textsc{RoBERTa} achieves an accuracy of 69.0\% on detecting fake articles published before 2019, but its accuracy drops to 51.9\% when testing on articles published after 2019. 

\paragraph{Multi-document reasoning} The rest of the incorrect detection is caused by the lack of multi-document reasoning ability. For instance, a news article\footnote{\url{https://tinyurl.com/multi-doc}} wrongly associates Hillary Clinton with a flawed immigration policy of the former government, and strengthens such a statement by referring to a Senate report and relevant news articles. However, the cited report does not mention Clinton, and the other news articles contain disinformation. To correctly detect this piece of disinformation, detectors should be capable of reasoning across multiple documents.

\section{Qualitative Examples of Generated Articles}

In \Cref{tab:qualitative_comparison}, we show a comparison of generated articles given the same input data across different generative methods. Our approach produces articles with a small fraction of inaccurate information, which matches a property of human-written fake news discussed in \Cref{sec:intro}.

\section{Appeal to Authority Details}

\label{apx:app_auth_details}
To recap, we first gather a list of authorities $Z$ for each article from Wikidata and the corresponding context. The best \textit{appeal to authority} sequence $s^*$ is selected with the lowest perplexity $s^* = \argmin_{s_i} \textrm{PPL}(s_i) $ where $s_i$ denotes the generated sequence using $z_i$ as the authority. However, this process results in every sequence $s^*$ containing the substring ``confirms that'', which makes it trivial for detectors to classify these generated documents as fake by simply detecting such substrings. Therefore, we devise an algorithm to diversify the templates so that these generated articles are not easily detectable. 

 First, we define a set of verbs $V$ that can be swapped with ``confirms'': $V=\{$\textit{said}, \textit{concluded}, \textit{confirmed}, \textit{emphasized}, \textit{stated}, \textit{argued}$\}$. Then, we diversify the generated structure of the generated sentence $s^*$ by reordering the subject, verb, and object. Next, we swap the verb with a another verb from $V$. Finally, to diversify the context, we append a preposition from the preposition set $PP =\{$\textit{on}, \textit{at}, \textit{in}$\}$ to the output of the previous step, and then feed the sequence to \textsc{BART} to generate the context. An example of this process is provided in \Cref{tab:appeal_to_authority_process}.

\section{Intermediate Pre-training Details}
\label{apx:ipt}
For domain adaptation, we perform intermediate pre-training (IPT) on the CNN/DM dataset, a large summarization corpus containing more than 280K news articles from CNN and Daily Mail. The IPT objectives for disinformation generation and propaganda generation are mostly the same as described in previous sections, but with some minor changes due to different goals in the IPT phase. When performing IPT for disinformation generation, we remove $\mathcal{L}_{s}$ from the final loss function (\Cref{eq:final_loss}) as the goal for IPT is only to learn to generate coherent sentences. In addition, to create training samples for \textit{loaded language} IPT, we gather all the appearances of adjectives pointing to a noun or adverbs pointing to a verb via dependency parsing graphs without considering whether the samples contain \textit{loaded} terms since the goal here is to enable \textsc{BART} to identify where to insert which adjectives or adverbs.
\begin{table}[t]
    \small
    \centering
    {
    \begin{tabular}{lcc}
        \toprule
        
        \textbf{Detector} & Dev Acc. (\%) & Test Acc. (\%)  \\ \midrule
        \textsc{HDSF} & 52.4 ($\pm$0.6) & 50.6 ($\pm$2.4) \\
        \textsc{Bert}  & 57.7 ($\pm$1.0) & 58.0 ($\pm$1.2)\\
        \textsc{Grover}  & 60.3 ($\pm$5.8) & 63.3 ($\pm$5.0)\\

        \textsc{RoBERTa} & \textbf{70.5} ($\pm$0.3) & \textbf{69.8} ($\pm$1.1)\\

        \bottomrule
    \end{tabular}
    }
    \vspace{-2mm}    
    \caption{Evaluation of various detectors on the \datasetname~ development and test set. We report the mean and standard deviation of four runs.} 
    \vspace{-5mm}
    \label{tab:propanews_test_results}
    
\end{table}
\begin{table*}[t]
    \setlength{\tabcolsep}{2pt}
    \small
    \centering
    
    {
    \begin{tabular}{p{0.1\linewidth} p{0.85\linewidth}}
        \toprule
        \textbf{Step} & \textbf{Generated Sequence}\\
        \midrule
        
        1 & \hlc{lightyellow}{Panmure Gordon analyst Peter Hitchens confirmed that ``} \hlc{lightblue}{the US government is likely to agree to reduce its estimate of the size of the spill, which would cut BP fines ''.} \\
        \midrule
        
        2 & `` The US government is likely to agree to reduce its estimate of the size of the spill, which would cut BP fines, '' Panmure Gordon analyst Peter Hitchens confirmed.\\
        
        \midrule
        3 & `` The US government is likely to agree to reduce its estimate of the size of the spill, which would cut BP fines, '' Panmure Gordon analyst Peter Hitchens said.\\
        \midrule
        
        4 & \hlc{lightyellow}{`` The US government is likely to agree to reduce its estimate of the size of the spill, which would cut BP fines, '' Panmure Gordon analyst Peter Hitchens said in} \hlc{lightblue}{a conference.} \\
        \bottomrule
    \end{tabular}
    }
    
    \caption{An illustration of how appeal to authority is performed. In step 1, we generate a statement using \textsc{BART} with the prefix ``Panmure Gordon analyst Peter Hitchens confirmed that `` ''. In step 2, we move the subject and verb to the back of the sentence to diversify the sentence structure. In step 3, we swap the verb with another verb from the verb set $V$. In step 4, we append a preposition \textit{in} to the sequence in step 3 and use the resulting sequence as prefix to \textsc{BART}'s decoder to generate the rest of the context. For step 1 and step 4, we mark the prefix sequence to the decoder in \hlc{lightyellow}{yellow}, and the generated sequence in \hlc{lightblue}{blue}. To increase the diversity of the generated sequences, step 2 to step 4 are each performed 50\% of the time.}
    \label{tab:appeal_to_authority_process}
    
\end{table*}

\section{Benchmarking Detectors} The performance of various detectors on the \datasetname~ dataset is shown in \Cref{tab:propanews_test_results}. We find that \textsc{RoBERTa} and \textsc{Grover} demonstrate advantages over \textsc{Bert}. This could be explained by the fact that \textsc{RoBERTa} and \textsc{Grover} are pre-trained on news domain corpora, whereas \textsc{Bert} has no access to such domains during pre-training. In addition, we find that \textsc{HDSF} performs much worse than the other three models. This reflects that large-scale pre-training of language models brings more benefit to detection performance than explicit modeling of discourse-level features.

\section{Human Validation Details}
\label{apx:annotation_interface}
In this section, we describe the details of human validation where AMT workers are tasked to validate whether the generated sentences contain inaccurate information. We recruit AMT workers from the United States and Canada. To ensure the annotation quality, only workers who have an acceptance rate greater than 95\% and have more than 100 accepted HITs in the past are allowed to work on our annotation task. This greatly reduce the chances of collecting annotations from scammers. Each HIT was designed such that the annotators are rewarded \$12-\$15 per hour, which complies with the ethical research standards outlined by AMT \cite{Salehi2015We}. In each HIT, the annotators are presented an article with the generated part marked in boldface. The questions and guidelines are illustrated as follows. (Note that we only use the annotators' response for Q1 to validate our generated data. The annotations for the other questions will be used for future research.)

\textbf{Q1:} Is the generated text in boldface \textbf{Accurate} or \textbf{Inaccurate}? (If you cannot find any supporting evidence, please select \textbf{Inaccurate}.) Note that:
A statement (in quotation marks) made by a person is only accurate if this person actually made the exact same statement. If the statement in quotation marks is just a paraphrase of what the person actually said, then the statement is inaccurate.
\begin{itemize}
  \item[-] \textbf{Inaccurate}: Any false information presented in the generated text makes it inaccurate.
  \item[-] \textbf{Accurate}: All the information in the generated text must be accurate.
\end{itemize}

\textbf{Q2:} Enter the URL of the news article you found that supports your decision in the previous response in the below box. Put down ``from context" if the evidence can be found in the context.

\textbf{Q3:} Does the generated text in boldface delivers the same sentiment as the rest of the article?
\begin{itemize}
  \item[-] \textbf{False}: The sentiment of the generated text is NOT the same as the rest of the article.
  \item[-] \textbf{True}: The sentiment of the generated text is the same as the rest of the article.
\end{itemize}

\textbf{Q4:} Is the discourse of the generated text in boldface consistent with the rest of the article?
\begin{itemize}
  \item[-] \textbf{False}: The discourse of the generated text is NOT consistent with the rest of the article. 
  \item[-] \textbf{True}: The discourse of the generated text is consistent with the rest of the article.
\end{itemize}
 
\textbf{Q5:} If there is any grammatical error or inconsistent discourse, please rewrite the correct the generated text and put it in the below box. Just put down the corrected generated text in bold is enough. For example, ``Harry is a boy. He likes go to school.'' Please put in ``He likes to go to school.'' in the box below.

\section{Statistics of the Evaluation Datasets}
In \Cref{tab:dataset_stats}, we show the statistics of the two evaluation datasets used in our experiments. The reported numbers are not the same as listed in the original papers \cite{nguyen2020fang, shu2018fakenewsnet} since some of the articles are no longer accessible via the provided URLs.
\label{apx:dataset_stats}
\begin{table}[h]
    \small
    \centering
    {
    \begin{tabular}{lccc}
        \toprule
        
        \textbf{Dataset}  & \textbf{\# Real} & \textbf{\# Fake }\\
         \midrule
        \textsc{Snopes} & 430 & 280  \\
        \textsc{PolitiFact} & 517 & 369\\

        \bottomrule
    \end{tabular}
    }
    
    \caption{Statistics of the two evaluation datasets, \textsc{Snopes} and \textsc{PolitiFact}.}
    \label{tab:dataset_stats}
    
\end{table}

\section{Detector Implementation Details}
\label{apx:implementation_details}
For \textsc{BERT} adn \textsc{RoBERTa} experiments, we use AdamW \cite{DBLP:conf/iclr/LoshchilovH19} as the optimizer with a batch size of 2 and gradient accumulation steps of 8. We set the learning rate and weight decay to 5e-5 and 1e-5 for the parameters that have been pre-trained, and 1e-3 and 1e-3 for other parameters. For experiments on the \textsc{Grover} detector, we follow the original detection setting. \textsc{Grover} is trained using Adam \cite{DBLP:journals/corr/KingmaB14} with a learning rate of 2e-5 and a batch size of 64. Similarly, we follow the original recipe to train \textsc{HDSF}, which is optimized with Adam with a learning rate of 1e-2. All detectors are fine-tuned for at most 20 epochs where the best model is determined by the accuracy on the development set.

All experiments are conducted on a Ubuntu 18.04 machine with NVIDIA Tesla V100. We use PyTorch 1.10.0 and Transformers 4.3.0 for constructing all models and loading pre-trained weights, except for \textsc{Grover}, which operates on Tensorflow 1.13.1. The training time for \textsc{BERT} and \textsc{RoBERTa}, which contains around 340M parameters, is around 2-3 hours, while the training time for \textsc{Grover}, which contains 355M parameters, is around 1 hour. 

\section{Human Evaluation Details}
\label{apx:human_eval_details}
In this section, we describe the survey we deliver to AMT workers for evaluating the quality of the generated articles. Annotators are presented a generated article and asked to answer a few questions regarding the quality of it. \textbf{Q2} is only applicable for evaluating generated articles from \datasetname~, in which we show the sentence that contains propaganda. The low, medium, and high ratings in the response correspond to 1, 2, and 3 scores described in \Cref{subsec:results}. The questions and guidelines are illustrated as follows:

\textbf{Q1:} How plausible do you think of the article above?

\begin{itemize}
  \item[-] \textbf{Low}: It likely contains inaccurate information.
  \item[-] \textbf{Medium}: Not sure.
  \item[-] \textbf{High}: It unlikely contain inaccurate information.
\end{itemize}

\textbf{Q2:} How much does this sentence in the article affects your decision on the previous answer?

\begin{itemize}
  \item[-] \textbf{Low}: This sentence does not affect my answer for the previous question.
  \item[-] \textbf{Medium}: This sentence somehow affect my answer for the previous question.
  \item[-] \textbf{High}: This sentence largely affects my answer for the previous question.
\end{itemize}

\begin{figure}[h]
    \centering
    \includegraphics[width=0.9\linewidth]{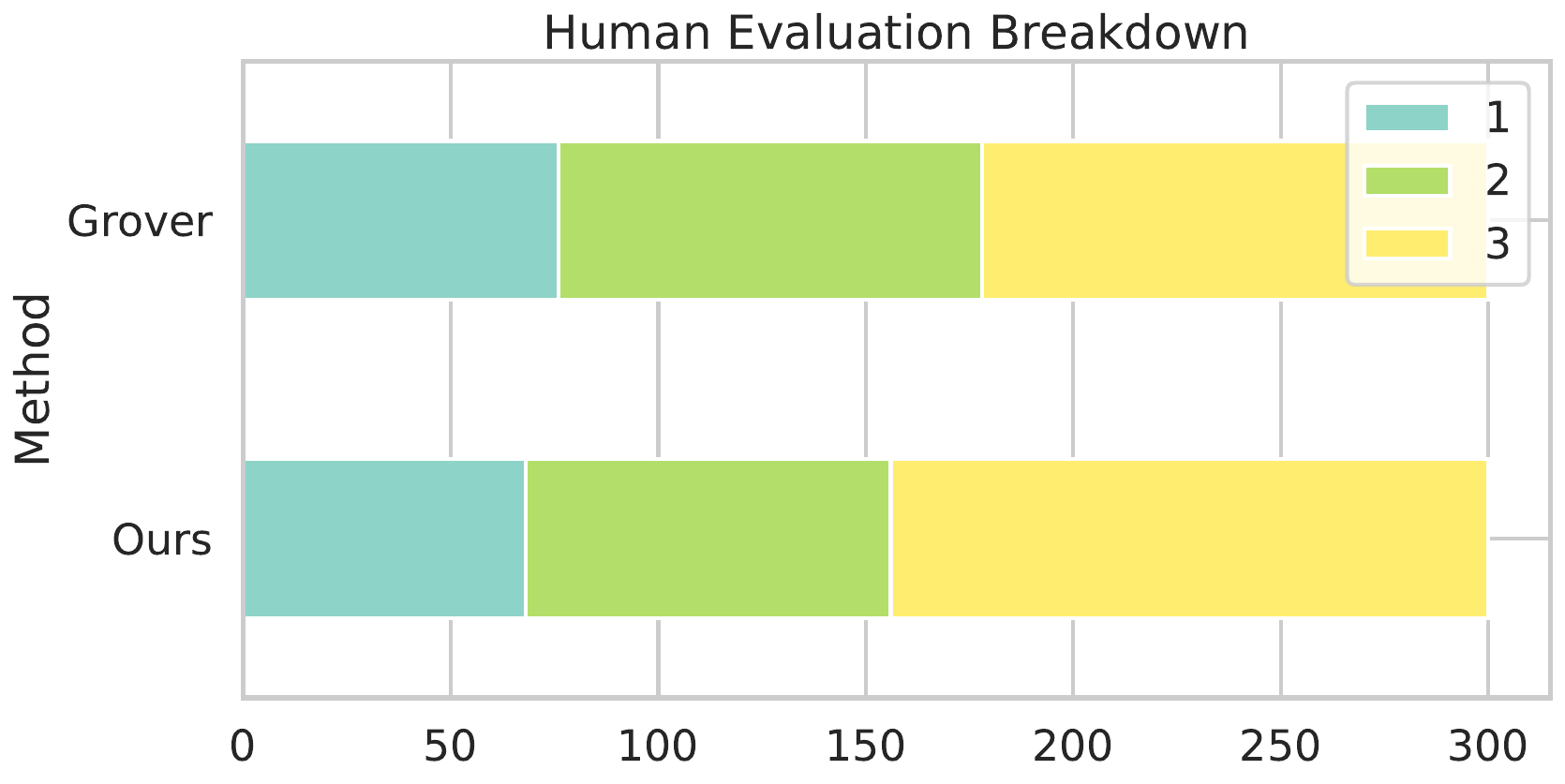}
    \caption{Breakdown scores of our human evaluation. The x-axis denotes the counts of evaluators votes for a score, while the y-axis denotes different methods.}
    \label{fig:human_eval_breakdown}
\end{figure}

The score distribution for \textbf{Q1} is shown in \Cref{fig:human_eval_breakdown}. We demonstrate that our approach produces higher quality fake news compared to \textsc{Grover-gen}.

\section{Scientific Artifacts}
We re-purposed the \textit{Timeline17} dataset \citep{Tran2013LeveragingLT} to produce our \datasetname~ dataset. 

Among the software we use, SpaCy (``en\_core\_web\_md'') is licensed under the MIT License, Transformers and Stanza are licensed under the Apache License 2.0, and PyTorch is released under the Modified BSD License. In terms of models, both BART and the NLI model we used are licensed under the MIT License.

In summary, all artifacts are produced with softwares and models released under open-source licenses. We will release the \datasetname~ dataset as well as the detectors with licenses compatible with the software discussed above such that they are consistent with their intended use.

\begin{table*}[t]
    \small
    \centering
    {
    \begin{tabular}{p{0.15\linewidth} p{0.8\linewidth}}
        \toprule
        \textbf{Method} & \textbf{Generated Article} \\
        \midrule
        
        \textsc{Grover-gen} & \hlc{lightorange}{Libya to restart three oilfields on Sept 1 : minister BENGHAZI , Libya , Aug 28 ( Reuters ) - Libya will restart two eastern oil fields in mid-September and resume crude shipment from Tobruk by the end of the month , SNCM deputy oil minister Habib Hummed said on Sunday . He said the west of Sirte was being restored and a third field in the eastern province of Brega was about to restart operations . `` Production at the Brega field has been increased by two to four percent and will resume its production on the end of the month , '' he said at a political conference . `` The second phase of production at the eastern field of Sirte came to full production by the end of the month and restarting the production from Brega field will begin in September , '' he said . He said production at the west of Sirte and east of Tobruk had already reached 95,000 barrels per day ( bpd ) on their main export terminals .} Libya , awash with millions of dollars , most of it believed to have been siphoned off through crude theft , is pumping around 290,000 bpd in total . The first fuel terminals in Libya have reopened and are loaded with diesel and petrol and tankers from Europe are ploughing in cargo to get through a key smuggling route from Derna to the Mediterranean . \hlc{lightorange}{Libya was sharply divided last year when armed tribesmen struck crude export terminals along its border , denying it the lucrative gas exports it was using to service southern EU countries . Libya 's southern oil terminal was hit in September by militant attacks . ( Writing by Ulf Laessing ; Editing by David Holmes )} \\
        \midrule
        \textsc{FactGen} & \hlc{lightorange}{Shares Augmented and stored pipelines replaced by light pipes .Disputed concerns over biggest damage to Libya since 2003 . Despite concerns oil could be leaking . 'Russia says production is safe and it is ready to resume production . 'Greece could use more sulfur oil by 2020 . 'Energy ministers say it could be used to power two-thirds of the world economy . 'Saudi had been the biggest oil producer in the world before meltdown . ''French to continue to contain crisis . ''Russia on Libyan crisis : ` we are facing a serious and continuing public health crisis that is needed ' ''But not in a vacuum of oil ' ''World affairs . ''Greece : ` we need to be doing more to combat damage to the energy system . ' '' leftasing all oil from Libya crisis . ''} \\
        \midrule
        Ours (\textsc{PN-silver})&  Tue Aug 23 , 2011 4:05 pm EDT * Limited Libyan oil production could start soon * Condition of oil export terminals still unclear * Clearing pipes clogged by waxy crude could delay output By David Sheppard and Sarah Young NEW YORK/LONDON , Aug 23 -LRB- Reuters -RRB- - Limited Libyan oil production ought to resume quite quickly as most fields appear to have survived the civil war unscathed , international oil services companies say , but many are still waiting for more stability in the country before sending employees back in . In interviews with some of the biggest oil services players in pre-civil war Libya , most echoed the view that major oil fields had largely avoided serious damage during the six months of fighting , though some voiced concerns about damage to export terminals and pipelines . OPS International Chairman Gavin De Salis told Reuters Insider television that Libyan crude oil , prized for its high yield of valuable light products such as gasoline and for its low sulfur content , was quite waxy , which could clog up pipelines if they had been left unused for some time . `` There might be a little bit of effort unplugging pipelines , which is two to three months ' worth of effort before they can resume full production , '' De Salis said . `` But that will not affect all of the pipelines or all of the fields , so they can certainly start limited production quite quickly . '' Nilsson said contacts at Libya 's rebel oil firm Arabian Gulf Oil Company -LRB- AGOCO -RRB- informed him there had been little damage to the oilfields in the east of the country during the six-month power struggle . `` We have n't been able to work at the oilfields during the civil war as it has not been safe , but I think within a couple of weeks we could be back to almost normal , '' Nilsson said by telephone from his office in Stockholm . `` The oil income is essential to Libya and the new government so they will want to bring it back online as soon as possible . '' Nilsson said they had several Swedish , Indian and Sudanese employees who had stayed in the country during the civil war , but total staff numbers in the country were down from around 250-300 . \hlc{lightorange}{Nilsson said there was still a lot of work to be done in the country .} \hlc{lightblue}{De Salis said that `` a lot of damage '' had been done to Libya 's oil infrastructure , including the destruction of some of the country 's main oil export terminals , but he said it was too early to estimate the full extent of the damage .} DAMAGE Oil firm 's who supported the rebel government during the civil war are expected to win the lion 's share of contracts to help relaunch the Libyan oil industry , which before the war produced some 1.6 million barrels per day of crude ... \\
        
        \bottomrule
    \end{tabular}
    }
    
    \caption{A qualitative comparison between generated articles from different approaches. The texts marked in \hlc{lightorange}{orange} indicate disinformation, and the texts in \hlc{lightblue}{blue} denote propaganda. We see that other approaches generate a large amount of inaccurate information, which contrasts with a property of human-written fake news mentioned in \Cref{sec:intro}. We also note that the article generated using \textsc{FactGen} appear to be low-quality. This is likely caused by the fact that the checkpoints reported in the paper were not released and we train \textsc{FactGen} from scratch by closely following the recipe described in \citet{Shu_Li_Ding_Liu_2021}. It is possible that some details of the training process of \textsc{FactGen} were missing from the paper, and hence the low generation quality. }
    \label{tab:qualitative_comparison}
    
\end{table*}
\end{document}